\pdfoutput=1
\documentclass{article}

\usepackage[preprint,nonatbib]{paper}

\usepackage[utf8]{inputenc} 
\usepackage[T1]{fontenc}    
\usepackage{url}            
\usepackage{booktabs}       
\usepackage{amsfonts}       
\usepackage{nicefrac}       
\usepackage{microtype}      
\usepackage{xcolor}         
\usepackage{amsmath}
\usepackage{amssymb}
\usepackage{graphicx}
\usepackage{booktabs}
\usepackage{xfrac}
\usepackage{color}
\usepackage{colortbl}
\usepackage{sidecap}
\usepackage{enumitem}
\usepackage{adjustbox}
\usepackage{romannum}
\usepackage{multirow}
\usepackage{wrapfig}
\usepackage{caption}
\usepackage{subcaption}
\usepackage{tikz,lipsum}
\usepackage{scalerel}
\usepackage{ifnextok}
\usepackage[most]{tcolorbox}
\definecolor{grey}{RGB}{128,138,135}
\definecolor{darkgrey}{RGB}{96,96,96}
\definecolor{citecolor}{HTML}{0071bc}
\definecolor{shadecolor}{HTML}{EFEFEF}
\usepackage[colorlinks, linkcolor=red, colorlinks, anchorcolor=blue, citecolor=citecolor, pagebackref=True]{hyperref}
\usepackage{cleveref}

\newcommand\extrafootertext[1]{%
    \bgroup
    \renewcommand\thefootnote{\fnsymbol{footnote}}%
    \renewcommand\thempfootnote{\fnsymbol{mpfootnote}}%
    \footnotetext[0]{#1}%
    \egroup
}

\usepackage{xspace}

\definecolor{lavender}{HTML}{E5E2FB}
\definecolor{lightblue}{HTML}{CEEBF9}
\definecolor{lightyellow}{HTML}{F7D9AE}
\definecolor{lightred}{HTML}{EEDCDB}
\definecolor{green}{HTML}{3BCB41}
\definecolor{darkgreen}{HTML}{156C09}
\definecolor{purple}{HTML}{9903F0}

\usepackage{titlesec}
\titlespacing\section{0pt}{8pt plus 4pt minus 2pt}{0pt plus 2pt minus 2pt}
\titlespacing\subsection{0pt}{8pt plus 4pt minus 2pt}{0pt plus 2pt minus 2pt}
\titlespacing\subsubsection{0pt}{8pt plus 4pt minus 2pt}{0pt plus 2pt minus 2pt}
\newcommand{\name}{BuboGPT\xspace}

\begin{document}


\title{
\begin{minipage}{.1\textwidth}
\includegraphics[width=\linewidth]{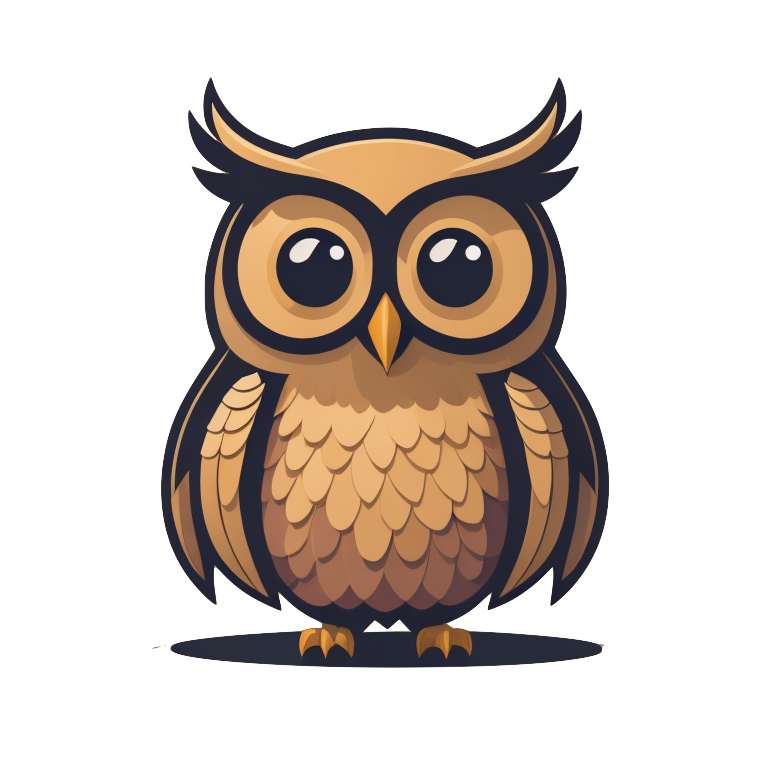} 
\end{minipage}%
\begin{minipage}{.75\textwidth}
\begin{center}
BuboGPT: Enabling Visual Grounding \\ in Multi-Modal LLMs
\end{center}
\end{minipage}

}

\author{%
  Yang Zhao$^*$, Zhijie Lin$^*$, Daquan Zhou, Zilong Huang, Jiashi Feng, Bingyi Kang$^\dag$ \\
  \texttt{\{zhaoyang98, linzhijie11, daquanzhou, zilonghuang, jshfeng, bingyikang\}} \\ \texttt{@bytedance.com} \\
  $*$ Equal Contribution, $\dag$ Project Lead \\
}

\maketitle

\begin{abstract}
LLMs have demonstrated remarkable abilities at interacting with humans through language, especially with the usage of instruction-following data. Recent advancements in LLMs, such as MiniGPT-4, LLaVA, and X-LLM, further enlarge their abilities by incorporating multi-modal inputs, including image, video, and speech. 
Despite their effectiveness at generating precise and detailed language understanding of the given modality signal, these LLMs give up the ability to ground specific parts of inputs, thus only constructing a coarse-grained mapping. However, explicit and informative correspondence between text and other modalities will not only improve the user experience but also help to expand the application scenario of multi-modal LLMs.
Therefore, we propose \name, a multi-modal LLM with visual grounding that can perform cross-modal interaction between vision, audio and language, providing fine-grained understanding of visual objects and other given modalities. As a result, \name is able to point out the specific location of an object in the image, when it is generating response or description for that object. 
Our contributions are two-fold: 1) An
off-the-shelf visual grounding module based on SAM that extracts entities in a sentence and find corresponding masks in the image.  2) A two-stage training scheme and instruction dataset to endow joint text-image-audio understanding. Our experiments show that \name achieves impressive multi-modality understanding and visual grounding abilities during the interaction with human. It performs consistently well when provided by arbitrary modality combinations (either aligned or unaligned). Our code, model and dataset are available at \url{https://bubo-gpt.github.io}.




\end{abstract}

\section{Introduction}
The large language models~(LLMs) have made significant progress and demonstrated promising abilities in few-shot and zero-shot learning by leveraging instruct tuning~\cite{wei2021finetuned} on carefully curated datasets. To harness the potential of LLMs beyond just language, 
some recent studies~\cite{zhu2023minigpt,chen2023x,zhang2023video,zhang2023llama,liu2023visual,li2023otter,su2023pandagpt,luo2023valley,maaz2023video} successfully connect LLMs with more input signals (\textit{e.g.}, image, video, speech and audio), and build powerful multi-modal chatbots. However, these models often perform understanding without digging into the fine-grained relation between the visual objects and other given modalities.
For example, when an illustrative figure is given, a visually-enhanced LLM will generate a high-quality description with rich details, but in a black-box manner. Instead, an instructive teacher-bot is going to show its audience which part of the figure it is referring to and what is happening there. Such visual grounding abilities are intriguing to LLMs but previously under-explored in the literature.


In this paper, we propose \name, the first attempt to incorporate visual grounding into LLMs by relating visual objects with other modalities. Moreover, it is able to perform joint multi-modal understanding and chatting for text, vision and audio, which is achieved by learning a shared representation space that aligns well with pre-trained LLMs. 

To this end, we first build an off-the-shelf visual grounding pipeline based on SAM~\cite{kirillov2023segment} to explore the fine-grained relation between different visual objects and modalities. The pipeline is composed of three modules, namely, a \emph{tagging module}, a \emph{grounding module} and a \emph{entity-matching module}. The tagging module is a pre-trained modal~\cite{zhang2023recognize} that can generate multiple text tags/labels that are relevant to the input image. The SAM-based~\cite{kirillov2023segment} grounding module~\cite{liu2023grounding} further localize the semantic mask or box on the image for each tag/label. Then, the entity-matching module leverages the reasoning capabilities of LLMs to retrieve matched entities from tags and image descriptions. In this way, we connect visual objects and other modalities by using language as a bridge. 


Then, to unlock the multi-modal understanding ability for arbitrarily combined inputs, we employ a two-stage training scheme similar to Mini-GTP4~\cite{zhu2023minigpt}. More specifically, we use ImageBind~\cite{girdhar2023imagebind} as the audio encoder,  BLIP-2~\cite{li2023blip} as the vision encoder and Vicuna~\cite{chiang2023vicuna} as the LLM. In the first stage, we learn a Q-former to align vision or audio features with language on image or audio caption datasets respectively. In the second stage, we perform multi-modal instruct tuning on a high-quality instruction-following dataset. We observe that the construction of this dataset is crucial for the LLM to recognize whether a modality is provided and whether the input modalities are well matched with each other. Therefore, we devise a novel high-quality dataset, which is composed of four subsets: 1) vision instruction dataset; 2) audio instruction dataset; 3) sound localization dataset with positively paired image-audio examples; 4) image-audio captioning dataset with negative pairs. Note that by introducing the negative image-audio pairs for semantic reasoning, the \name can learn better multi-modal alignment and demonstrate stronger capabilities of joint understanding.



Our experiments show that \name achieves impressive visual grounding abilities during multi-modal chat, even when arbitrary combinations of multi-modal inputs are provided, whether matched or unmatched.  
We summarize our key contributions as follows:
\begin{itemize}
     \item We build a multi-modal LLM, \name for multi-modal understanding including image, audio and text by learning a common semantic space and further explore the fine-grained relation between different visual objects and different modalities.
    \item We construct a high-quality multi-modal instruction-tuning dataset including fine-grained audio descriptions and cross-modal sound localization, and introduce both positive and negative image-audio pairs for semantic matching to facilitate the cross-modal understanding.
\end{itemize}

\begin{figure}[t]
    \centering
    \includegraphics[width=1.0\textwidth]{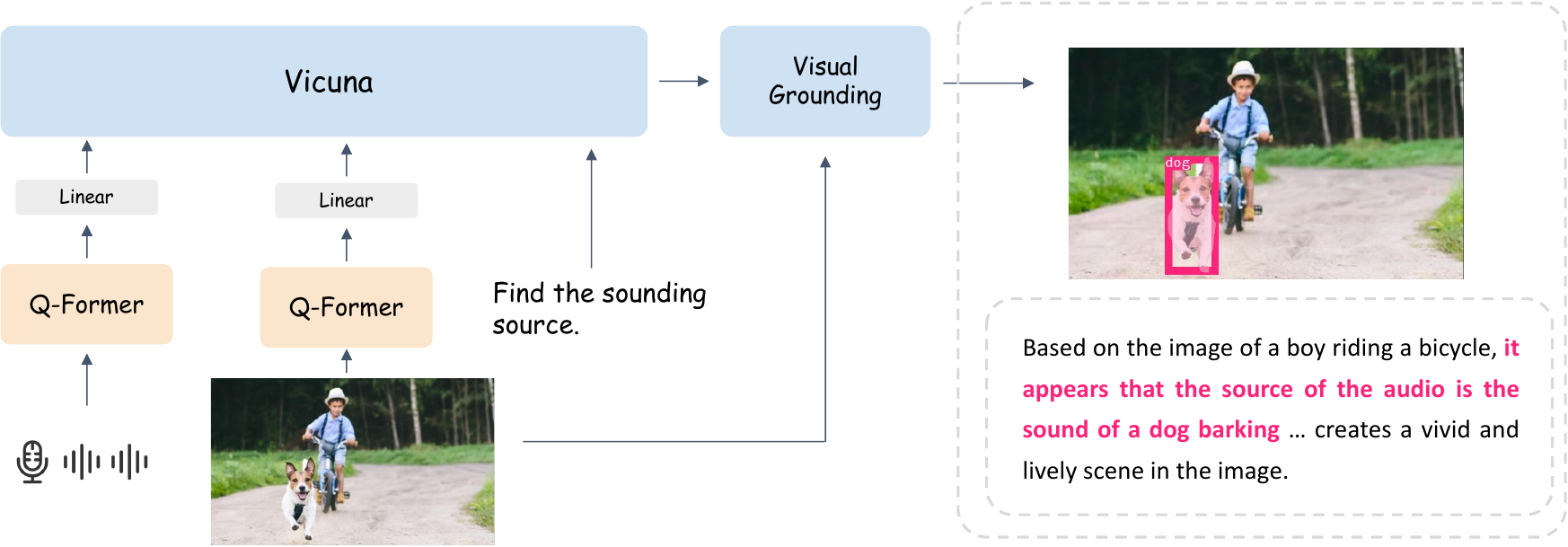}
    \caption{The overall framework of \name.}
    \label{fig:framework}
\end{figure}

\section{Related Work}

\textbf{Pre-trained LLMs in Mutli-modal Learning.} 
Due to the scaling up of training data and model size, large language models~\cite{openai2023gpt4,touvron2023llama,chowdhery2022palm,chiang2023vicuna} have demonstrated remarkable abilities across various linguistic tasks in a few-shot and zero-shot manner and also enabled conversational communication with humans. To leverage the powerful linguistic abilities of LLMs, some methods~\cite{wu2023visual,shen2023hugginggpt} propose to connect different accessedation models for multi-modal tasks by using LLMs as a dispatch scheduler. 

Based on high-quality multi-modal instruction-following data, recent end-to-end methods~\cite{zhu2023minigpt,chen2023x,zhang2023video,zhang2023llama,liu2023visual,li2023otter,su2023pandagpt,luo2023valley,maaz2023video} have been introduced to extend LLMs for multi-modal learning as well. Some works such as Mini-GPT4~\cite{zhu2023minigpt}, X-LLM~\cite{chen2023x} and Video-ChatGPT~\cite{maaz2023video} propose to align the input features of different modalities with pre-trained LLMs by learned visual encoder. Some works such as LLaMA-Adapter~\cite{zhang2023llama} and
Otter~\cite{li2023otter} insert learnable cross-attention layers into the pre-trained LLMs to incorporate multi-modalities knowledge. These prior methods mainly focus on tackling visual inputs~(e.g. videos and images)~\cite{zhu2023minigpt,zhang2023llama,liu2023visual,zhang2023video,luo2023valley,li2023otter} or ignoring the fine-grained relation between the visual objects and other given modalities~\cite{su2023pandagpt,chen2023x}. We further attempt to incorporate visual grounding into LLMs by relating visual objects with other modalities and propose to learn multi-modal alignment including image, audio and text in a common space.

\textbf{Multi-modal Instruction Tuning Dataset.}
To explore instruction tuning for multi-modal learning, \cite{xu2022multiinstruct} first introduces a multi-modal instruction tuning benchmark that is composed of 62 diverse multi-modal tasks in a unified seq-to-seq format. Mini-GPT4~\cite{zhu2023minigpt} curates an instruction following dataset by combining Conceptual Caption~\cite{sharma2018conceptual,changpinyo2021conceptual}, SBU~\cite{ordonez2011im2text} and LAION~\cite{schuhmann2021laion} with hand-designed prompt, while LLaVA~\cite{liu2023visual} proposes to use GPT-4~\cite{openai2023gpt4} to generate more detailed captions to expand COCO dataset~\cite{lin2014microsoft}. Otter~\cite{li2023otter} further builds a multi-modal in-context tuning dataset to facilitate the in-context learning capabilities of multi-modal LLMs. Further, we build a high-quality instruction tuning dataset including fine-grained audio description and introduce the negative image-audio pairs for semantic reasoning to enhance the reasoning capabilities of our model.

\section{Methods}
The overall framework of \name is presented in Figure~\ref{fig:framework}. As the Figure shown, we perform joint multi-modal understanding and chatting for text, vision and audio, which is achieved by learning a shared representation space that aligns well with pre-trained Vicuna~\cite{chiang2023vicuna}. We also build an off-the-shelf visual grounding pipeline to explore the fine-grained relation between different visual objects and modalities.

\subsection{Visual Grounding Pipeline}
\begin{figure}[t]
    \centering
    \includegraphics[width=1.0\textwidth]{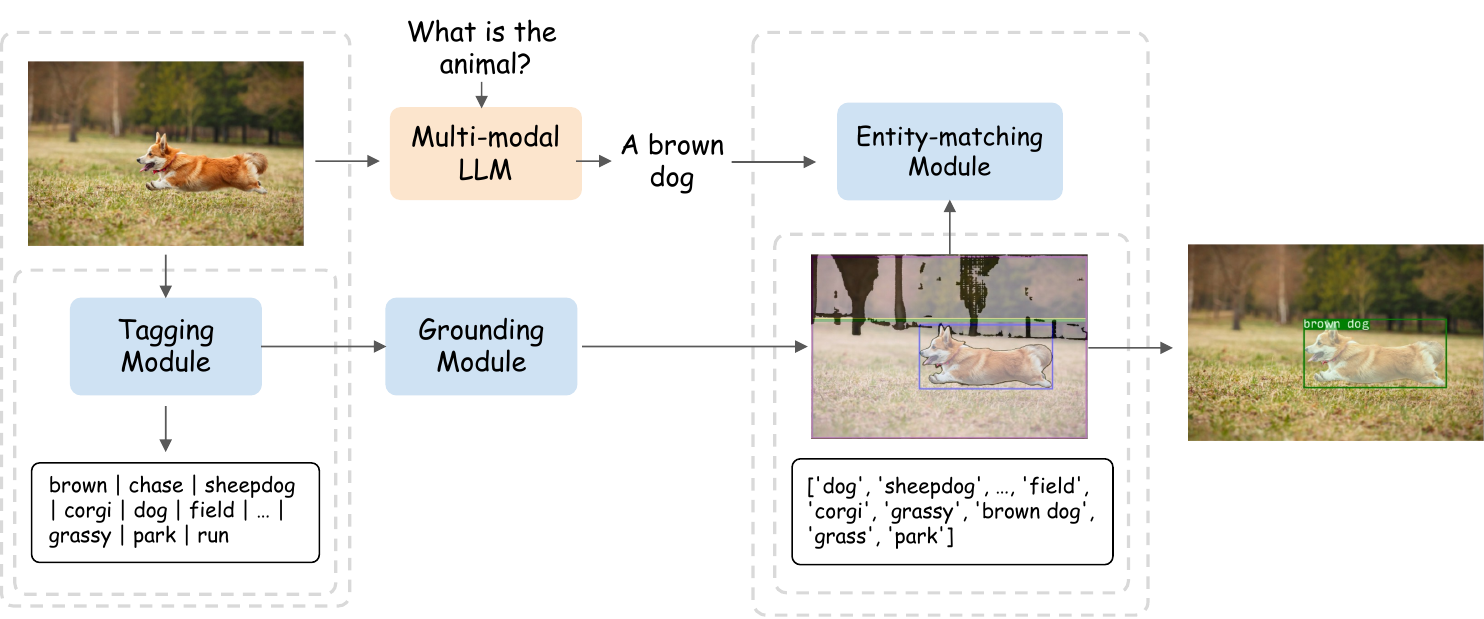}
    \caption{The pipeline of visual grounding that is composed of a tagging module, a grounding module and a entity-matching module.}
    \label{fig:grounding}
    \vspace{4mm}
\end{figure}

To explore the relation between different visual objects and input modalities, we further build the visual grounding pipeline, composed of a tagging module, a grounding module and a entity-matching module, as shown in Figure~\ref{fig:grounding}. Concretely, for a given image, we first use {Recognize Anything Model~(RAM)}~\cite{zhang2023recognize}, a strong model based on Swin-transformer~\cite{liu2021swin} for image tagging to generate relevant candidate tags, denoted as $\{t_1, t_2, ..., t_{n_t}\}$, where $t_i$ is the $i$-th semantic tag and $n_t$ is the number of detected tags. We then connect the tags with comma to form the prompt ``$t_1,t_2,...,t_{n_t}$'' and use the Grounding DINO~\cite{liu2023grounding}, a open-set object detection model with referring textual queries to identify the visual entities and the corresponding boxes relevant to the tags. Followed by {Segment Anything Model~(SAM)}~\cite{kirillov2023segment}, the boxes are taken as prompt to get fine-grained semantic masks.

With the tagging and grounding module, we then obtain all the visual entities and the corresponding grounding information, denoted as $\{(e_1, \boldsymbol{g_1}),(e_2, \boldsymbol{g_2}),...,(e_{n_e}, \boldsymbol{g_{n_e}})\}$, where $e_i,\boldsymbol{g_i}$ are separately the $i$-th visual entities and grounding information~(i.e. boxes and masks), $n_e$ is the number of entities. To model the relation between different visual entities and input modalities, we employ the text output $\boldsymbol{t}_o$ of our multi-modal LLM as the bridge and build a entity-matching module based on GPT-4 to retrieve the matching pairs. To construct the prompt template ``\textit{<List>$e_1, e_2, ..., e_{n_e}$</List>,<Text>$\boldsymbol{t}_o$</Text>}'', we utilize the powerful LLM for reasoning and retrieve the matching pairs, which reflects the relation between visual entities and input modalities.

\subsection{Multi-Modal LLM Training}
\label{sec:mm_training}
\name considers the interaction between three modalities, \textit{i.e.}, text, vision and audio. It aligns a vision encoder and an audio encoder with the LLM with a Q-former for each modality. More specifically, we utilize the visual encoder together with the pre-trained Q-Former in BLIP-2~\cite{li2023blip} and audio encoder in ImageBind~\cite{girdhar2023imagebind} for visual and audio perception. For joint understanding over multiple modalities, we employ Vicuna as the LLM. We use a linear projection layer to connect the modality Q-Former with the LLM. To effectively train such a model, we develop the following two-stage training scheme. The modality encoders and Vicuna model with be fixed throughout the training procedure.

\paragraph{Stage 1: Single-modal Pre-training} Similar to MiniGPT-4~\cite{zhu2023minigpt}, the first stage is designed to align the output of the linear projection layer to the word embedding space of the LLM. This is achieved by training the modality Q-Former and linear projection layer on a large number of modality-text paired data. For visual perception, we only train the projection layer for image captioning with the Q-Former from BLIP2 fixed. For audio understanding, we jointly train the Q-Former and the projection layer for audio captioning. There will not be any prompt used for both settings, the model just take the corresponding image or audio as input and predict the corresponding caption. 

\paragraph{Stage 2: Multi-Modal Instruct Tuning} This stage aims to equip the multi-modal LLM with the ability to understand human instructions such that it can generate proper responses based on the given modality signal. To this end, we curate a high-quality multi-modal instruction-following dataset, which contains image-text, audio-text and image-audio-text pairs. To make the model adapt to arbitrary combination of input modalities, we design a general prompt as: \textit{\#\#\#Human: <Vision><ModalityHere></Vision> <Audio><ModalityHere></Audio> \textcolor{olive}{<instruction>} \#\#\#Assistant:}. \textit{<Vision></Vision>} and \textit{<Audio></Audio>} are special identifiers for image and audio input. \textit{<ModalityHere>} is going to be replaced by a sequence of image or audio tokens before feeding into the LLM. \textit{\textcolor{olive}{<instruction>}} is the human instruction related to the input sigals for the LLM to assist on. We list a few examples for different combinations of input modalities in Tab.~\ref{tab:prompts}. We empirically accessed that when only positively paired image-audio data are included in this stage, the model always assumes the image and audio are related to each other even though random sampoles are used at test time. Therefore, we manually create some negative pairs and asking the LLM to tell what are they respectively. The experiments show that introducing such negative paired data is able to overcome this problem significantly. We leave the creation of datasets in the next section.

\begin{table}
\begin{tcolorbox}[colback=red!5!white,colframe=red!75!black]
\#\#\#Human: <Vision><ModalityHere></Vision> \textcolor{olive}{What is the image?} \#\#\#Assistant:

\#\#\#Human: <Audio><ModalityHere></Audio> \textcolor{olive}{Pay attention to the audio and describe what you notice.} \#\#\#Assistant:

\#\#\#Human: <Vision><ModalityHere></Vision> <Audio><ModalityHere></Audio> \textcolor{olive}{Please find the source that emits the given sound in this image.} \#\#\#Assistant:

\#\#\#Human: <Vision><ModalityHere></Vision> <Audio><ModalityHere></Audio> \textcolor{olive}{Are the audio and image related to each other? What are they?} \#\#\#Assistant:
\end{tcolorbox}
\caption{Instruction-following prompt examples for various input sources.}
\label{tab:prompts}
\end{table}

\section{Datasets}

\subsection{Pretraining Datasets}
Following MiniGPT-4~\cite{zhu2023minigpt}, we use a combined dataset of CC3M~\cite{sharma2018conceptual}, CC12M~\cite{changpinyo2021conceptual}, SBU~\cite{ouyang2022training}  and LAION~\cite{schuhmann2021laion} to train the visual projection layer, resulting in a total of 130 million image-text pairs. For audio, we mainly use the WaveCaps~\cite{mei2023wavcaps} dataset, which contains 403,050 audio clips with average duration of 67.59 seconds and average caption length of 7.8 words. It combines four datasets including FreeSound (262,300)~\cite{font2013freesound}, BBC Sound Effects (31,201)\footnote{\url{https://sound-effects.bbcrewind.co.uk/}}, SoundBible (1,231)\footnote{\url{https://soundbible.com/}} and AudioSet strongly-labelled subset (108,317)~\footnote{\url{https://research.google.com/audioset/download\_strong.html}}, and transform their raw-descriptions into captions with ChatGPT. 

\subsection{Instruction-Tuning Datasets}
\subsubsection{Image-Text Dataset}
\label{sec:image-text-data}
We employ two previously published datasets for visual instruct tuning. The first one is released by MiniGPT-4, which contains 3,439 high-quality text-image pairs. The second one provided by LLaVA~\cite{liu2023visual} is curated from 158K samples based on the COCO dataset, including three types of instructions, \textit{i.e.}, converstaions (58K), detailed description (23K) and complex reasnoning (77K). 

\subsubsection{Audio-Text Dataset}
\label{sec:audio-text-data}
When it comes to the field of audio understanding, we also need to conduct the instruction-tuning operation on the audio Q-former. However, unlike vision-language understanding, a severe need still exists for high-quality and well-organized instruction-tuning datasets in this field. To this end, we generate a series of expressive and descriptive data to facilitate this process. 

Specifically, we first investigate different kinds of existing audio caption datasets and select Clotho \cite{Drossos2019ClothoAA} as the original dataset to make the description extension. The reason can be explained in two folds. On the one hand, it has a moderate and acceptable scale to act as an instruction-tuning dataset, and the semantic range of audio is large enough. On the other hand, every audio has five short captions from different annotators, covering various possible scenes related to the audio and increasing the diversity of descriptions. 

After obtaining the original data, we need to rewrite the short captions into descriptive and imaginative paragraphs. Considering the extraordinary ability of GPT-4 in the field of few-shot learning, text generation, and complex reasoning, we utilize it to help us automatically assemble short captions into long descriptions to mitigate the reliance on human annotation. The final description is expected to cover all the related original captions. For example, given the series of captions \textit{[``A person is turning a map over and over.'', ``A person is very carefully wrapping a gift for someone else.'', ``A person is very carefully wrapping a gift for someone else.'', ``He sighed as he turned the pages of the book, stopping to scan the information.'', ``Papers are being turned, stopped, then turned again, and someone is breathing.'']}, the description paragraph is expected to be \textit{``A person is repeatedly flipping some papers. They might be reading a book, flipping through a map, or wrapping presents. Judging from the repeated flipping sounds, they are concentrating on repeating this action.''}. We design a task-related prompt and construct some few-shot examples like this to promote the in-context reasoning process. As a result, we collect a novel dataset \textit{Clotho-Detail} \footnote{\url{https://huggingface.co/datasets/magicr/BuboGPT/blob/main/Clotho-detail-annotation.json}} for instruction-tuning in audio understanding, which contains 3938 items and the average length of descriptions is 52.70 words. 

\subsubsection{Audio-Image-Text Dataset}
\label{sec:audio-image-text-data}
\textbf{Positive Set} In order to further empower our model with the comprehensive ability of multi-modal reasoning, we apply a group of audio-image pairs to help the model to understand the correspondence between the audio and its source. Among the existing audio-vision datasets, VGGSS \cite{Chen2021LocalizingVS} turns out to be a better choice in this process. It covers a wide range of sounding objects, and the audio only relates to a specific region in the corresponding image. Therefore, we retrieve all the data cases and use a group of fixed templates to wrap the corresponding class labels into natural sentence descriptions. As a result, we generate a total of 5,158 \textit{<audio, image, text>} pairs to act as the triple-modality instruction tuning dataset \footnote{\url{https://huggingface.co/datasets/magicr/BuboGPT/blob/main/vggss-instruction-tuning.json}}.

\textbf{Negative Set} As discussed in the method section (Sec.~\ref{sec:mm_training}), relying solely on the above dataset causes the LLM fail to recognize irrelevant audio-image pairs. Therefore, we construct negative \textit{<audio, image, text>} pairs such that \textit{<text>} gives independent descriptions for audio and image inputs. The audio is randomly sampled from the audio-text dataset presented in Sec.~\ref{sec:audio-text-data}, while the image is randomly sampled from the MiniGPT-4 dataset discussed in Sec.~\ref{sec:audio-text-data}. The text is constructed by concatenating the two captions that starts with ``The image'' and ``The audio''.

\section{Experiment Results}
In this section, we aim to answer the following two questions:  1) whether our \name is able to provide accurate and instructive visual grounding when the inputs contain images? 2) whether the modal is able to perceive arbitrary combinations of modalities and generate proper responses. 

\begin{figure*}[t]
  \centering
  \includegraphics[width=\linewidth]{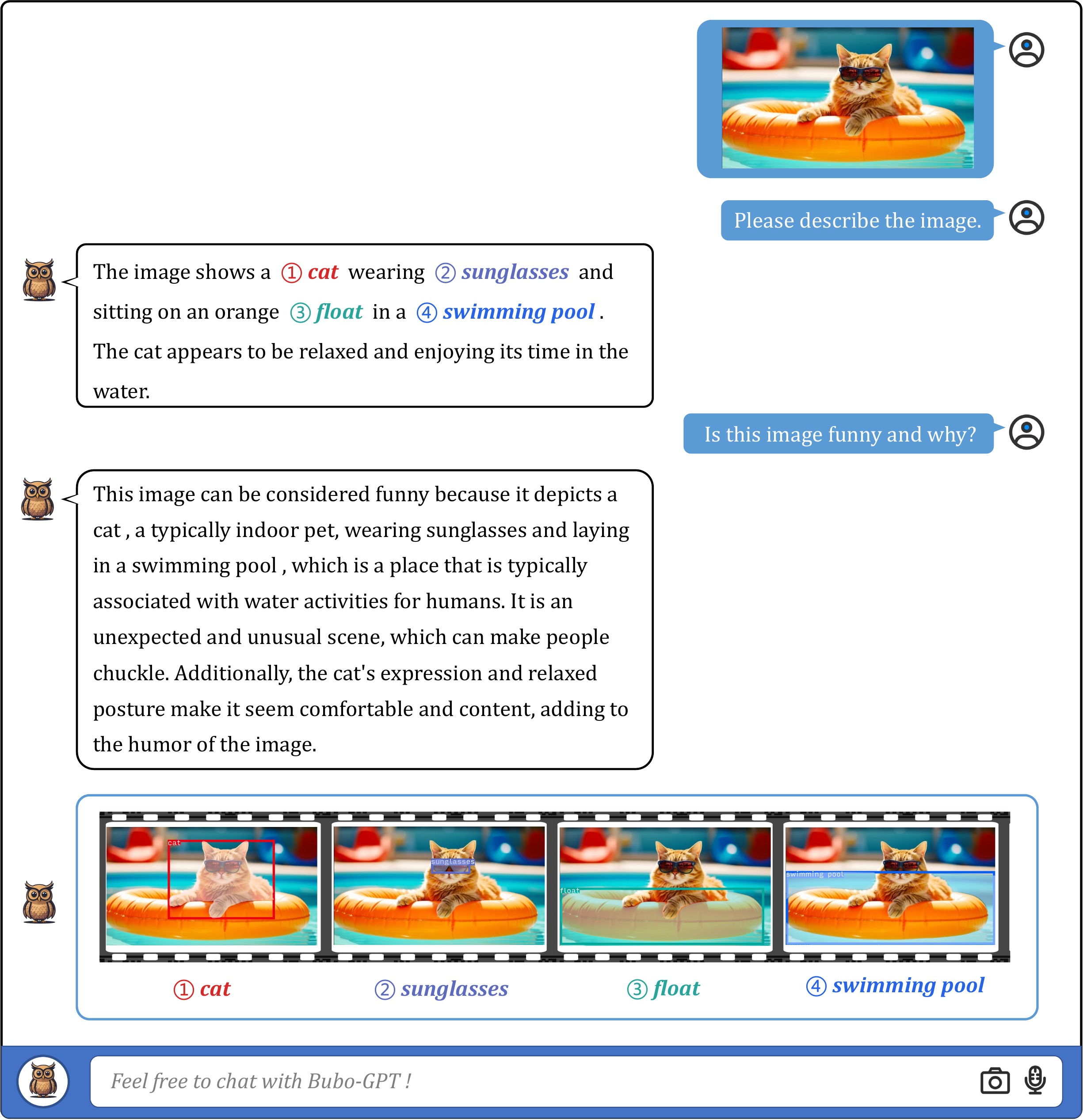}
  \caption{Example 1 of fine-grained visual understanding with grounding.}
  \label{figure:exp-pic-1}
\end{figure*}

\begin{figure*}[t]
  \centering
  \includegraphics[width=\linewidth]{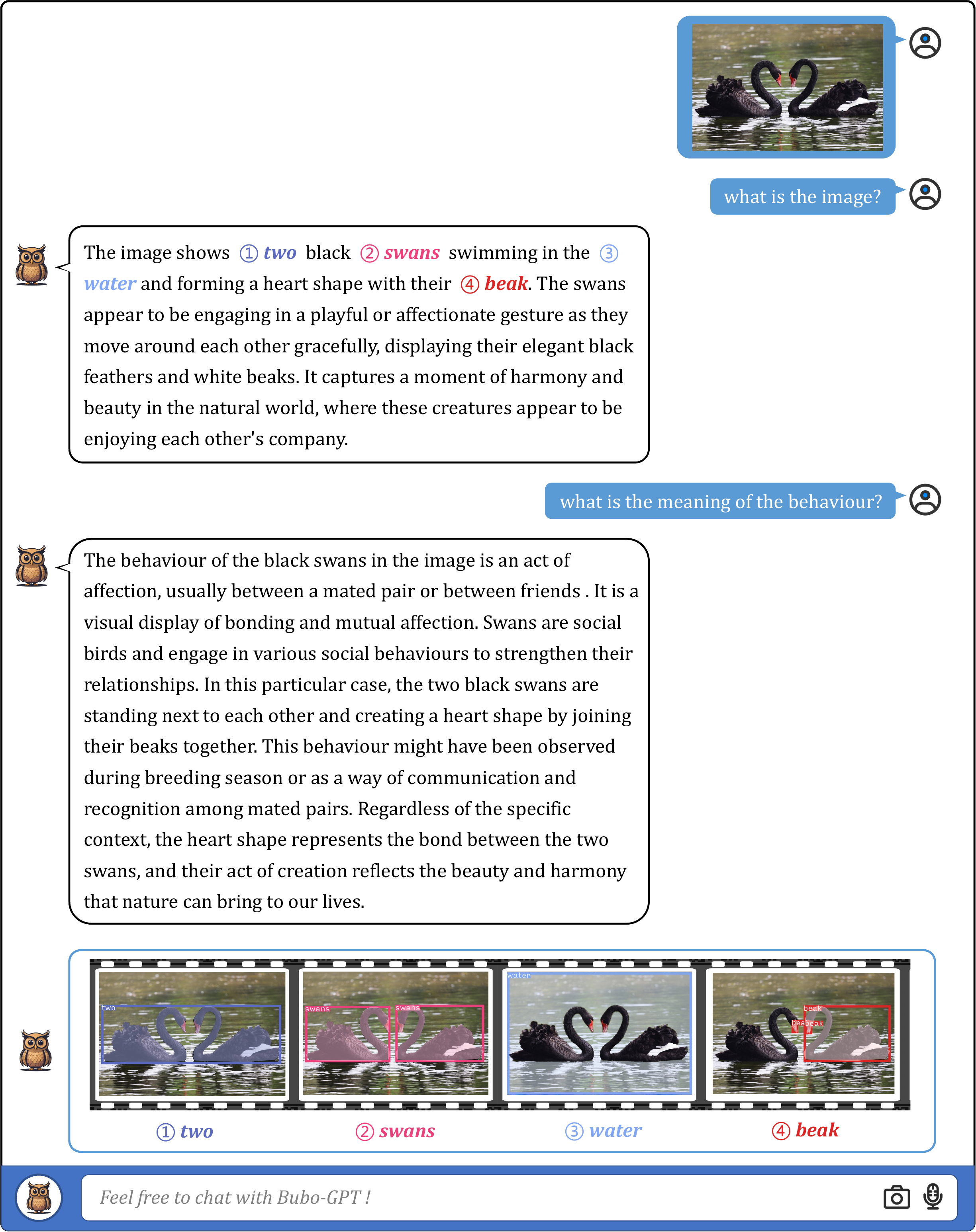}
  \caption{Example 2 of fine-grained visual understanding with grounding.}
  \label{figure:exp-pic-2}
\end{figure*}

\begin{figure*}[t]
  \centering
  \includegraphics[width=\linewidth]{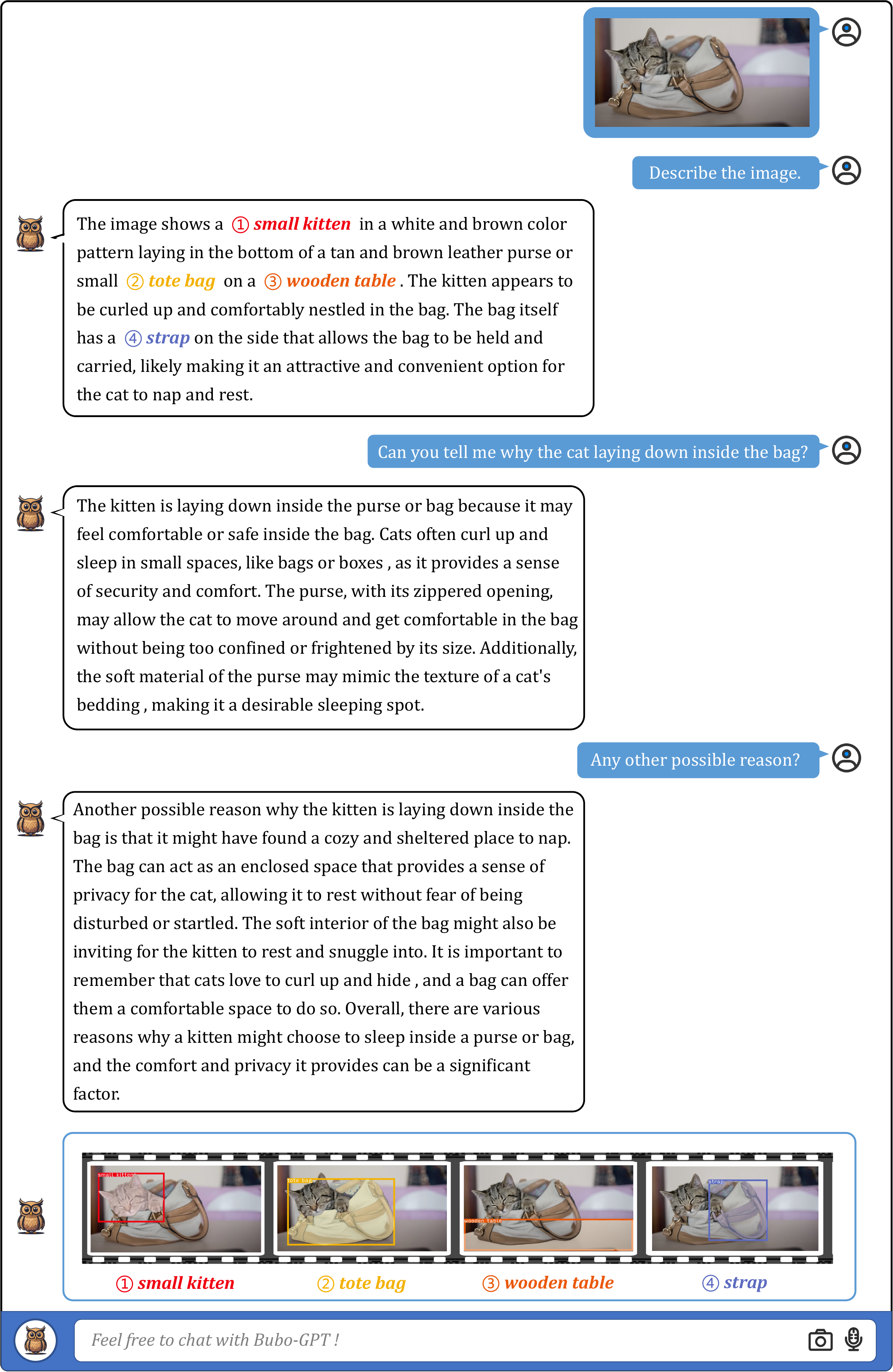}
  \caption{Example 3 of of fine-grained visual understanding with grounding.}
  \label{figure:exp-pic-3}
\end{figure*}

\begin{figure*}[t]
  \centering
  \includegraphics[width=\linewidth]{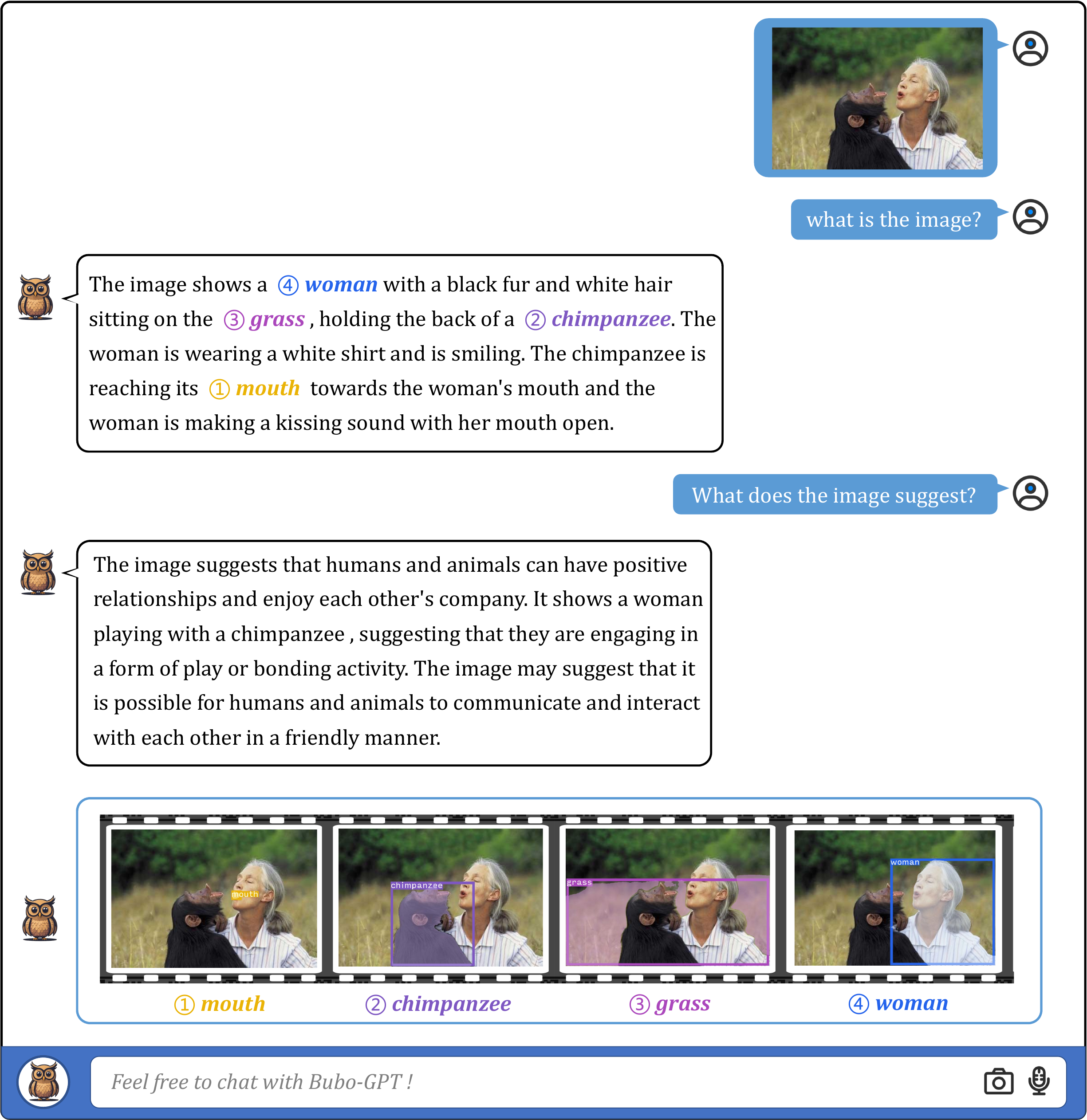}
  \caption{Example 4 of of fine-grained visual understanding with grounding.}
  \label{figure:exp-pic-4}
\end{figure*}

\begin{figure*}[t]
  \centering
  \includegraphics[width=\linewidth]{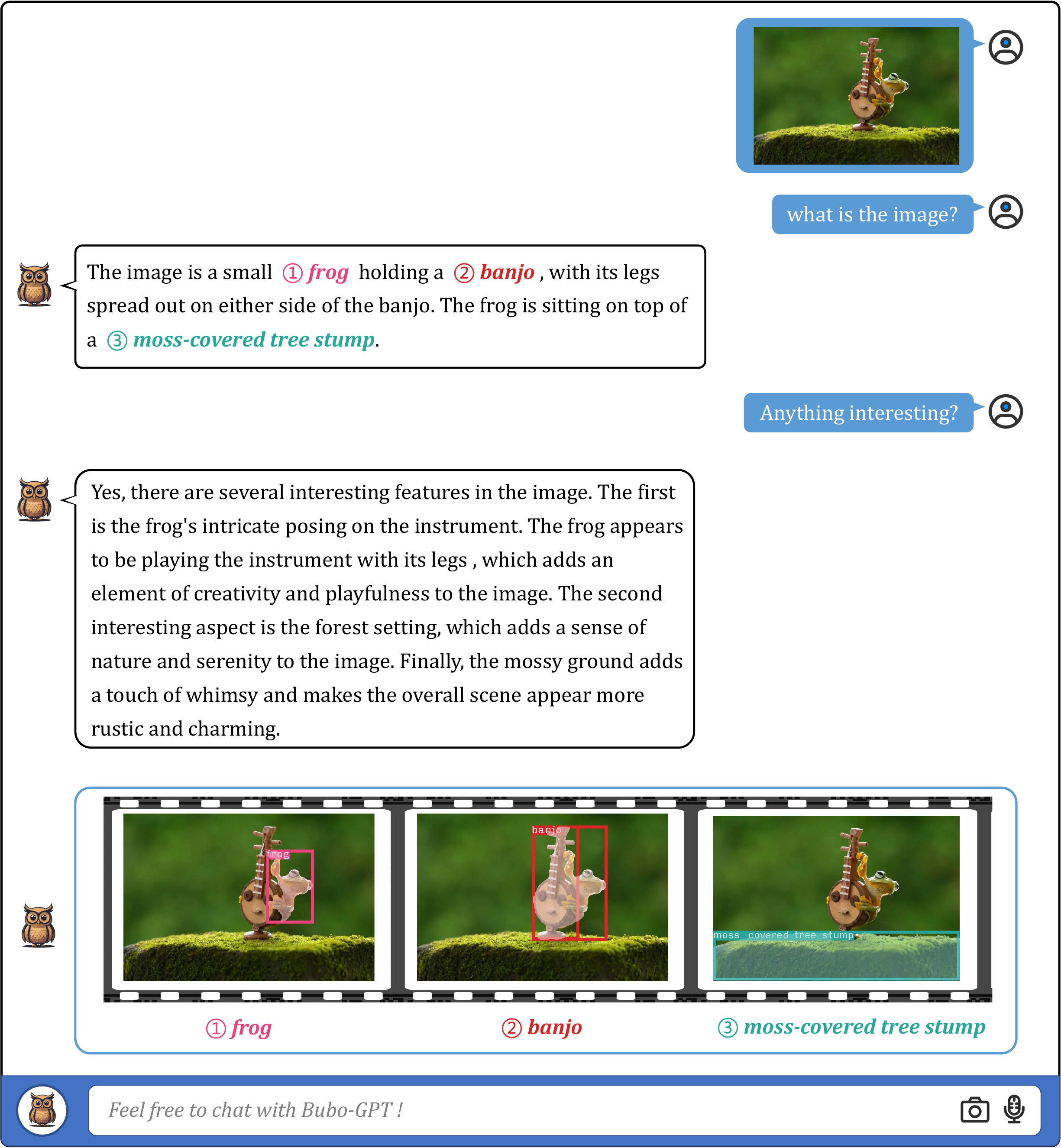}
  \caption{Example 5 of of fine-grained visual understanding with grounding.}
  \label{figure:exp-pic-5}
\end{figure*}

\begin{figure*}[t]
  \centering
  \includegraphics[width=\linewidth]{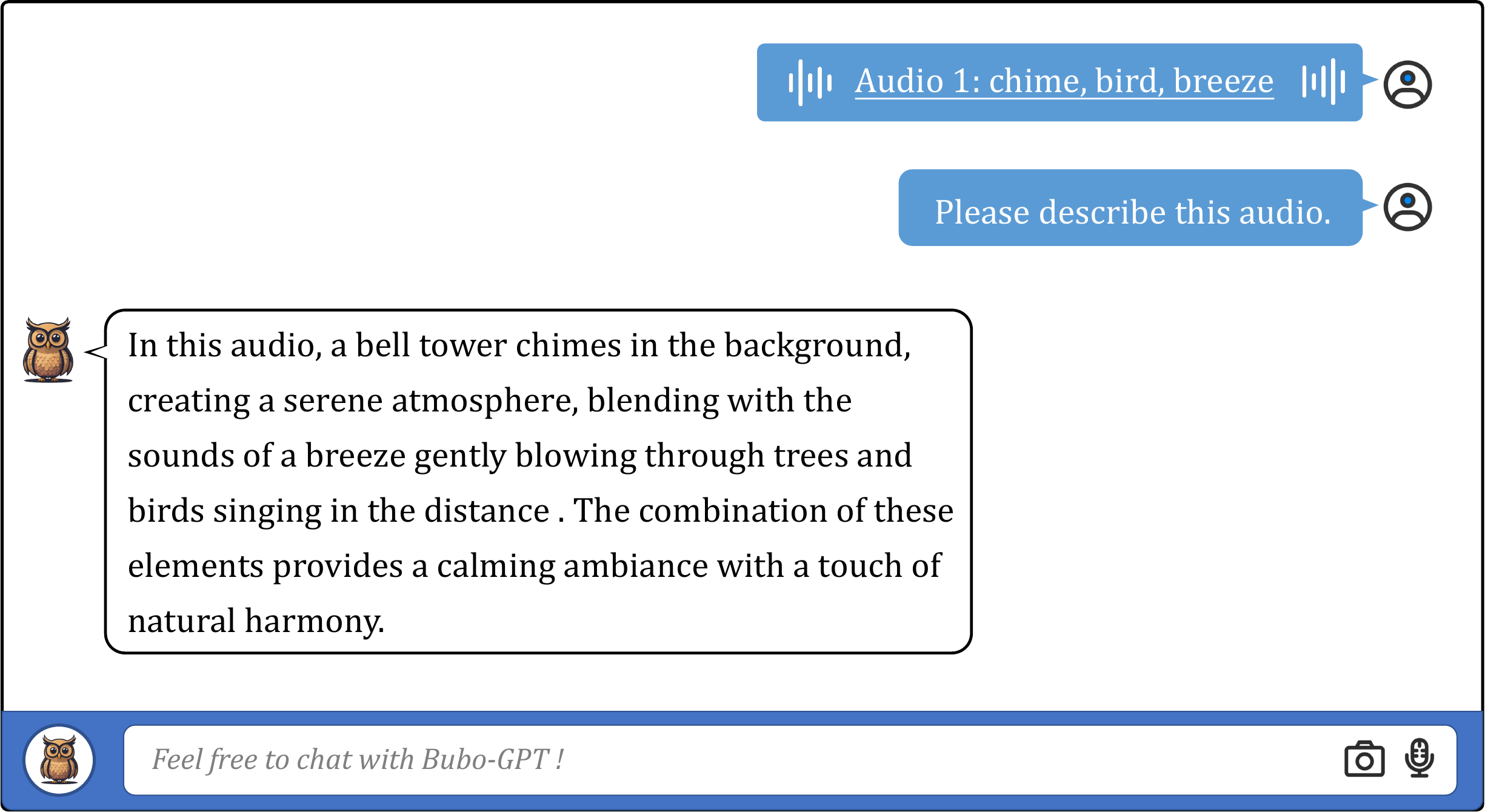}
  \caption{Example 1 of audio understanding. Audio 1 can be accessed via this \href{https://huggingface.co/datasets/magicr/BuboGPT/blob/main/Audios/Audio-1-chime-bird-breeze.wav}{link}.}
  \label{figure:exp-aud-1}
\end{figure*}

\begin{figure*}[t]
  \centering
  \includegraphics[width=\linewidth]{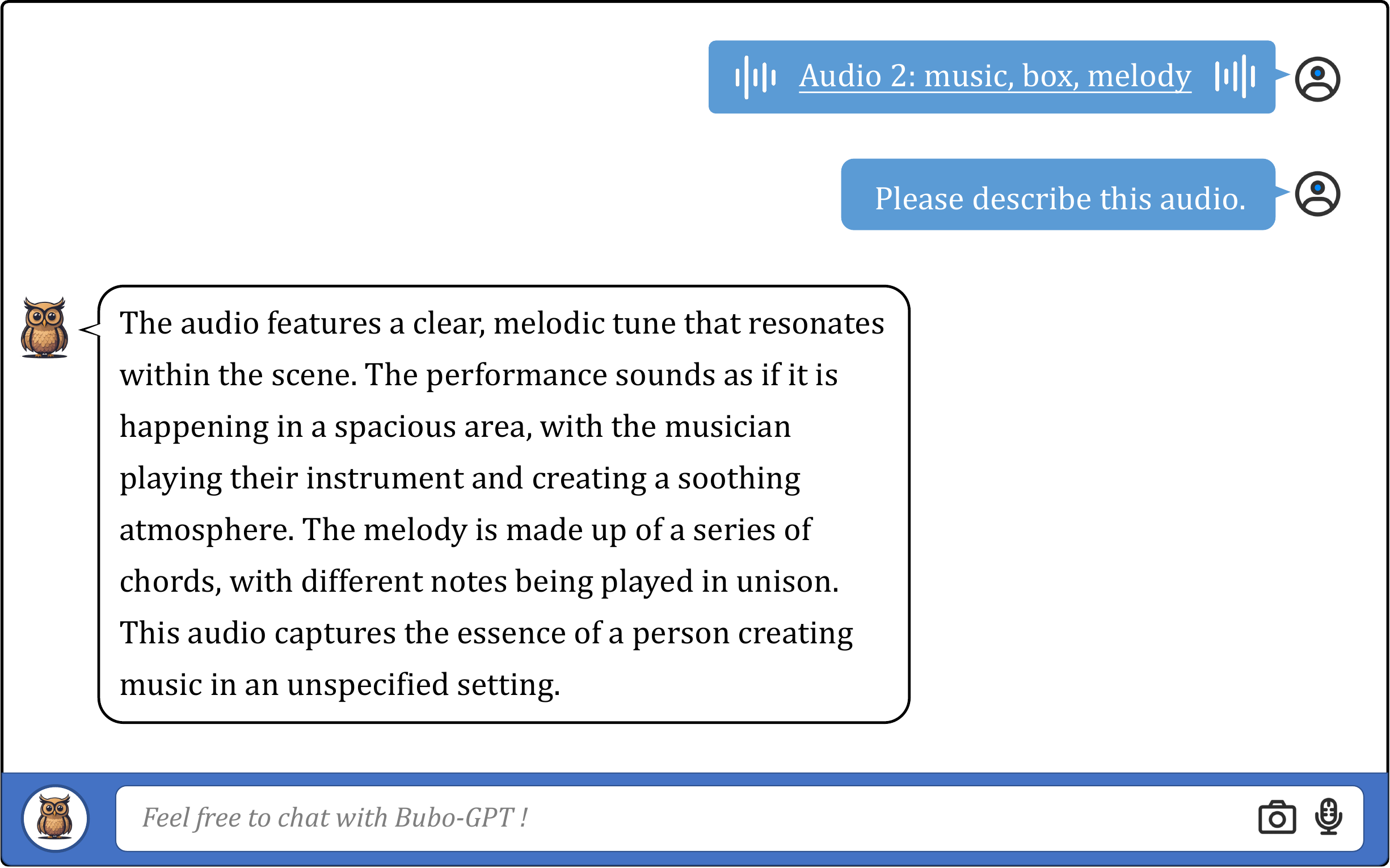}
  \caption{Example 2 of audio understanding. Audio 2 can be accessed via this \href{https://huggingface.co/datasets/magicr/BuboGPT/blob/main/Audios/Audio-2-music-box-melody.mp3}{link}.}
  \label{figure:exp-aud-2}
\end{figure*}

\begin{figure*}[t]
  \centering
  \includegraphics[width=\linewidth]{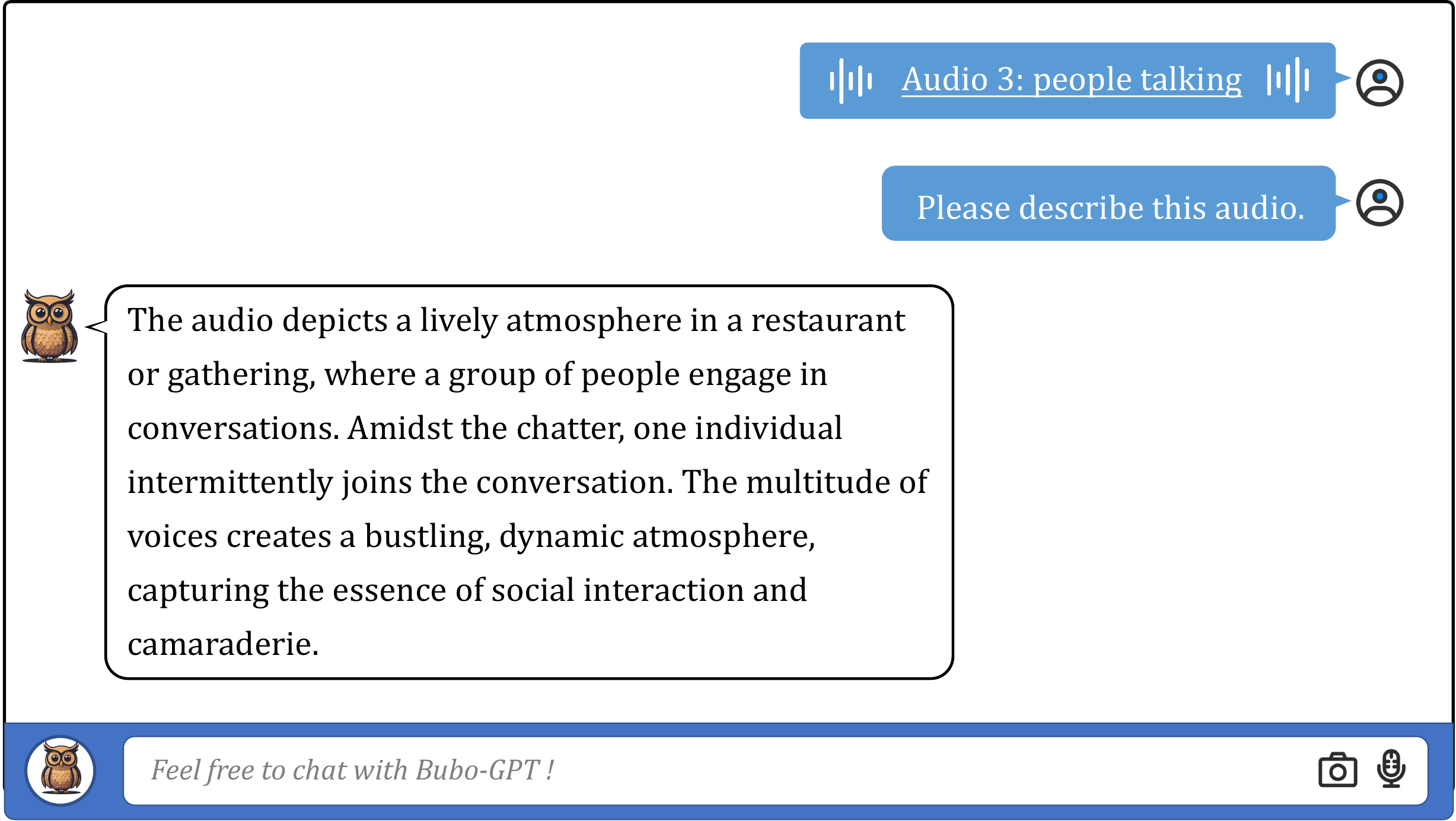}
  \caption{Example 3 of audio understanding. Audio 3 can be accessed via this \href{https://huggingface.co/datasets/magicr/BuboGPT/blob/main/Audios/Audio-3-People-Talking.flac}{link}.}
  \label{figure:exp-aud-3}
\end{figure*}

\begin{figure*}[t]
  \centering
  \includegraphics[width=\linewidth]{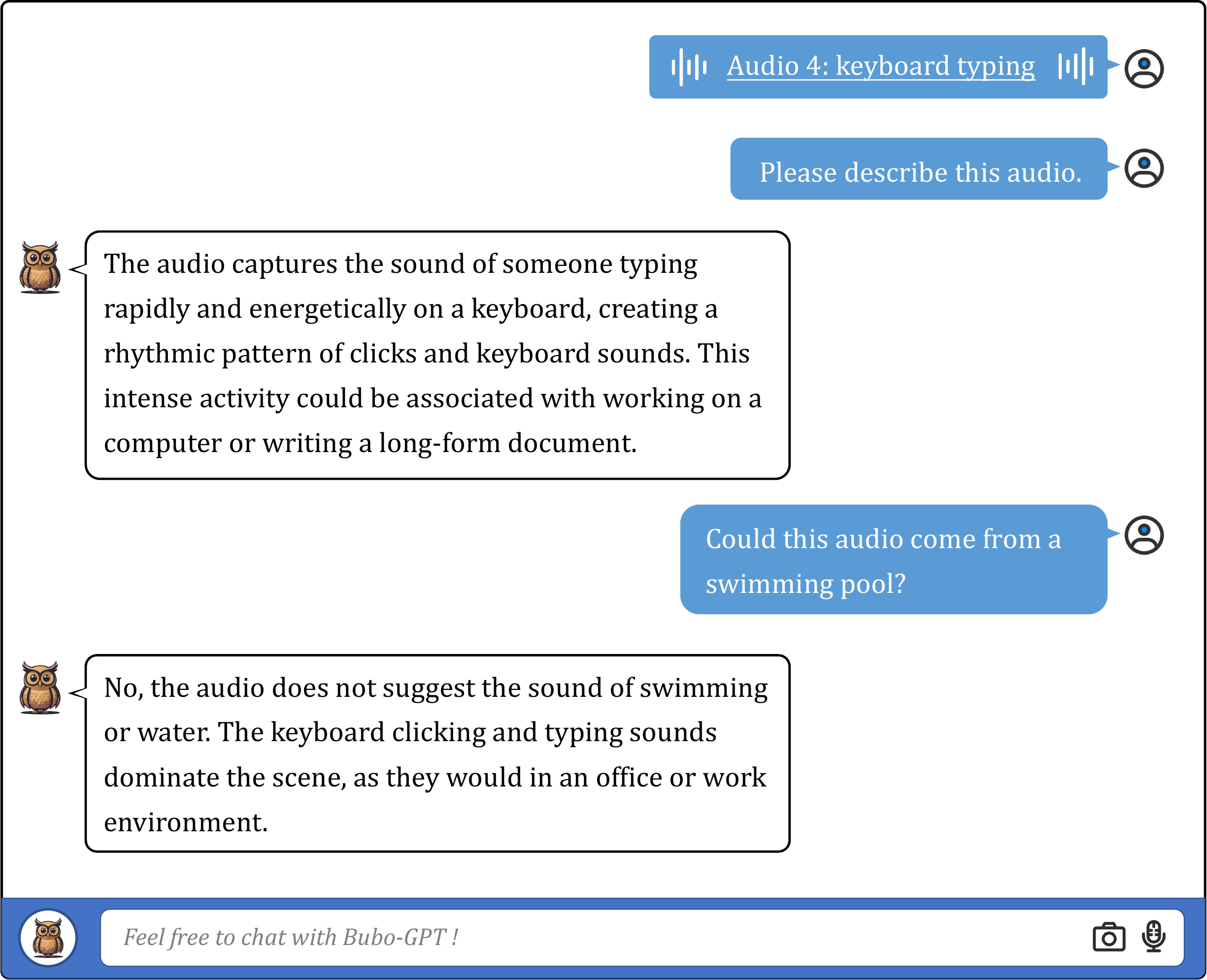}
  \caption{Example 4 of audio understanding. Audio 4 can be accessed via this \href{https://huggingface.co/datasets/magicr/BuboGPT/blob/main/Audios/Audio-4-keyboard-typing.mp3}{link}.}
  \label{figure:exp-aud-4}
\end{figure*}

\begin{figure*}[t]
  \centering
  \includegraphics[width=\linewidth]{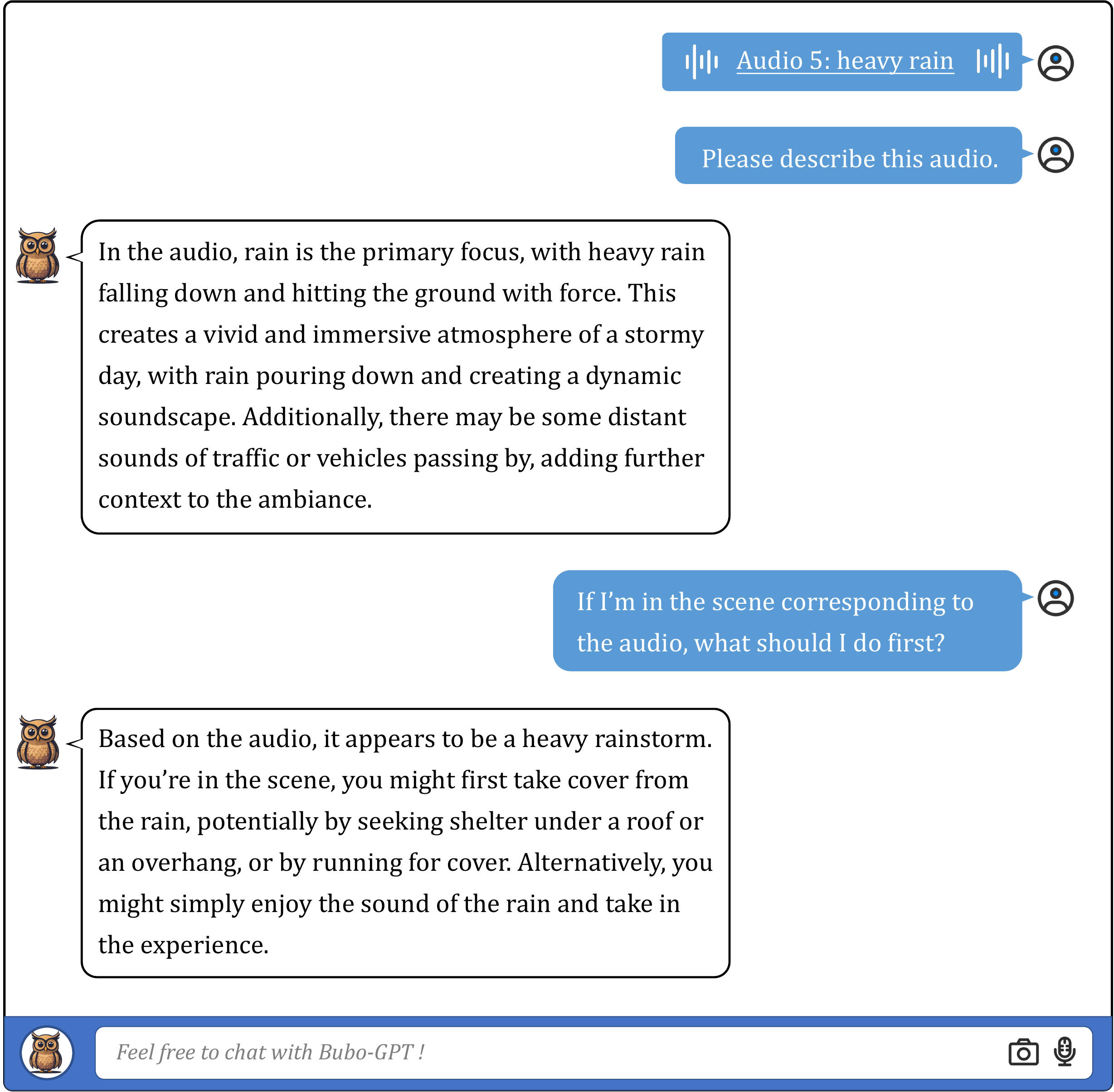}
  \caption{Example 5 of audio understanding. Audio 5 can be accessed via this 
 \href{https://huggingface.co/datasets/magicr/BuboGPT/blob/main/Audios/Audio-5-heavy-rain-sound.mp3}{link}.}
  \label{figure:exp-aud-5}
\end{figure*}

\begin{figure*}[t]
  \centering
  \includegraphics[width=\linewidth]{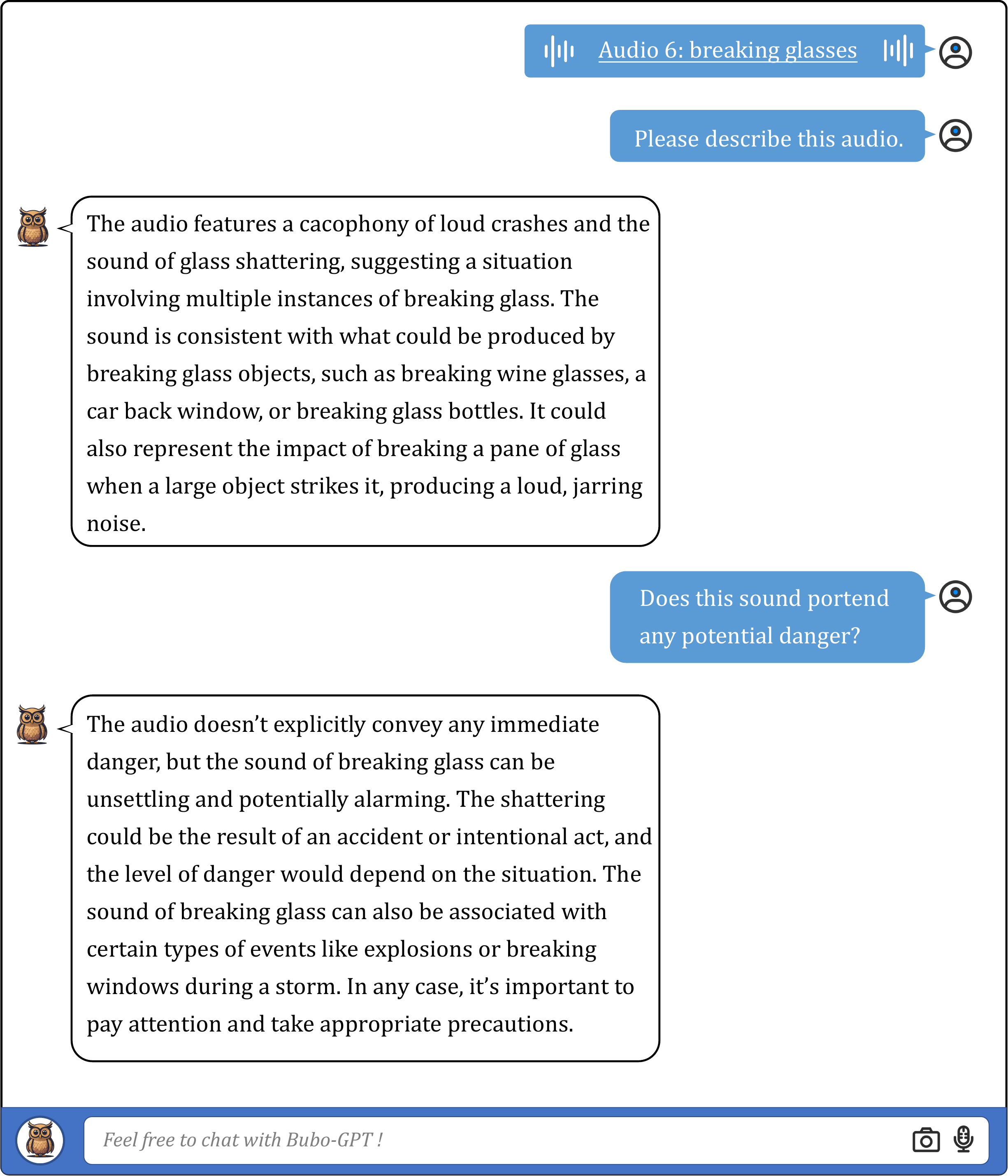}
  \caption{Example 6 of audio understanding. Audio 6 can be accessed via this \href{https://huggingface.co/datasets/magicr/BuboGPT/blob/main/Audios/Audio-6-breaking-glass.mp3}{link}.}
  \label{figure:exp-aud-6}
\end{figure*}

\begin{figure*}[t]
  \centering
  \includegraphics[width=\linewidth]{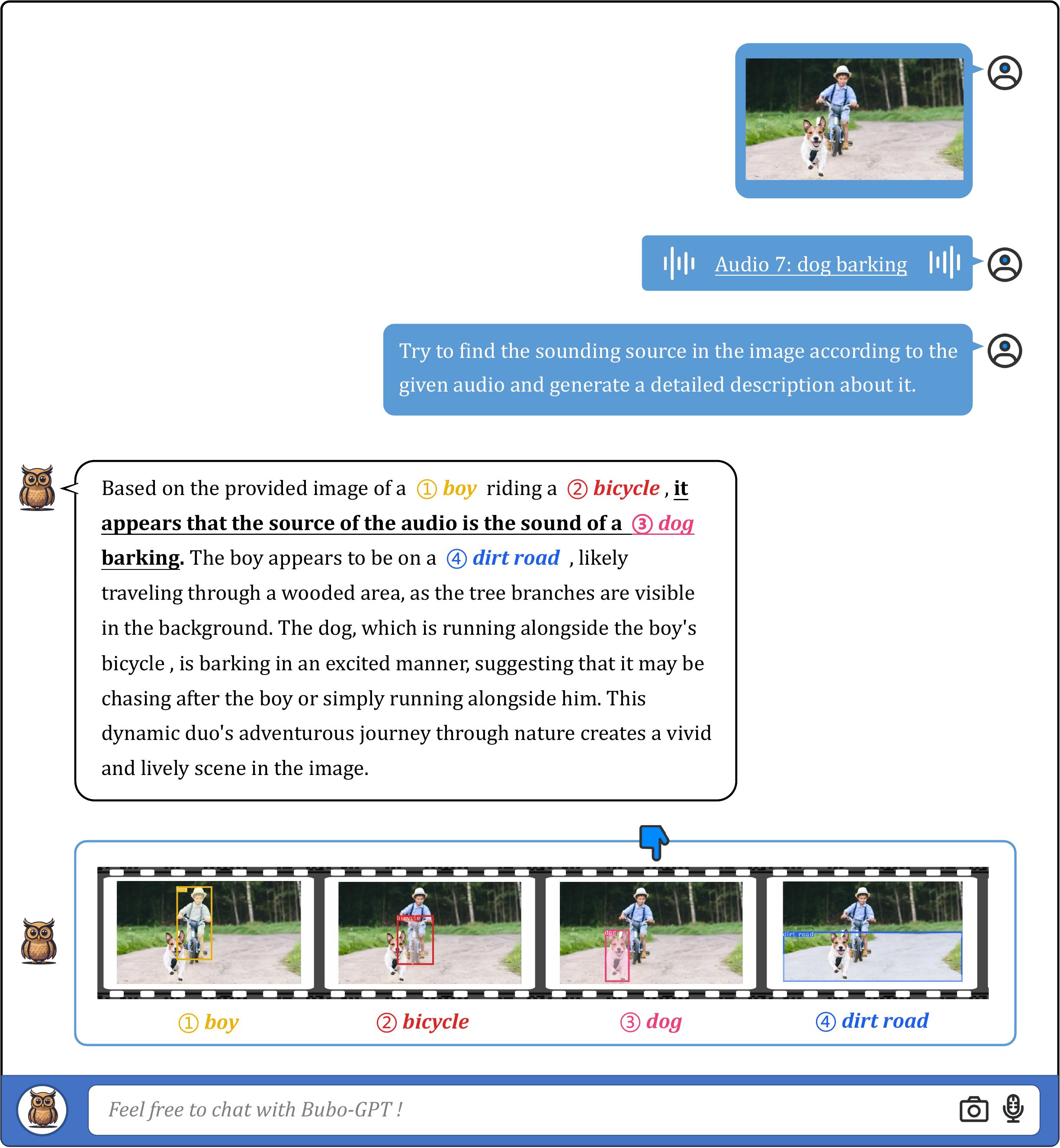}
  \caption{Example 1 of aligned audio-image understanding. Audio 7 can be accessed via this \href{https://huggingface.co/datasets/magicr/BuboGPT/blob/main/Audios/Audio-7-dork-bark.wav}{link}.}
  \label{figure:exp-pic-aud-1}
\end{figure*}

\begin{figure*}[t]
  \centering
  \includegraphics[width=\linewidth]{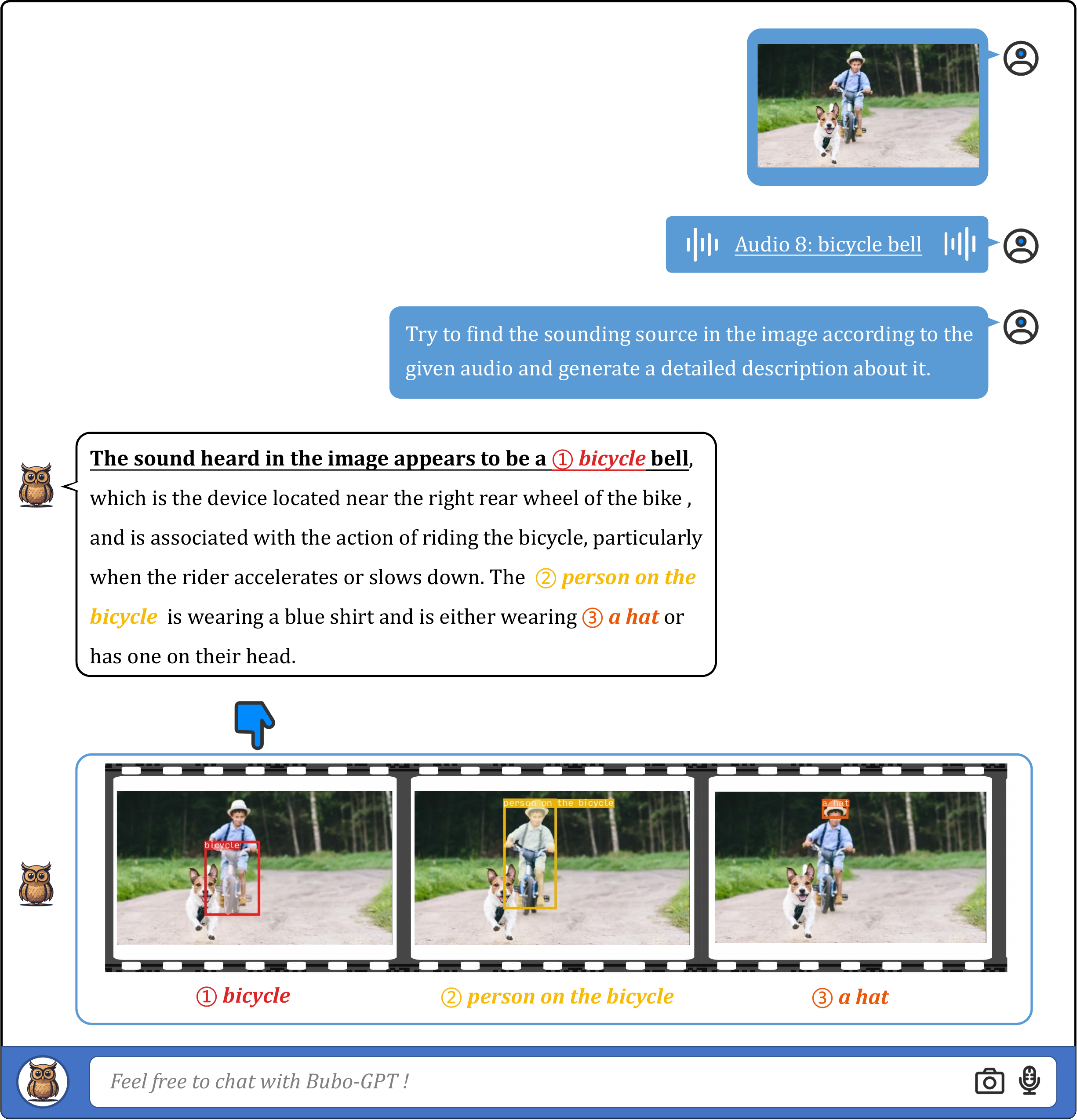}
  \caption{Example 2 of aligned audio-image understanding. Audio 8 can be accessed via this \href{https://huggingface.co/datasets/magicr/BuboGPT/blob/main/Audios/Audio-8-bicycle_bell.wav}{link}.}
  \label{figure:exp-pic-aud-2}
\end{figure*}

\begin{figure*}[t]
  \centering
  \includegraphics[width=\linewidth]{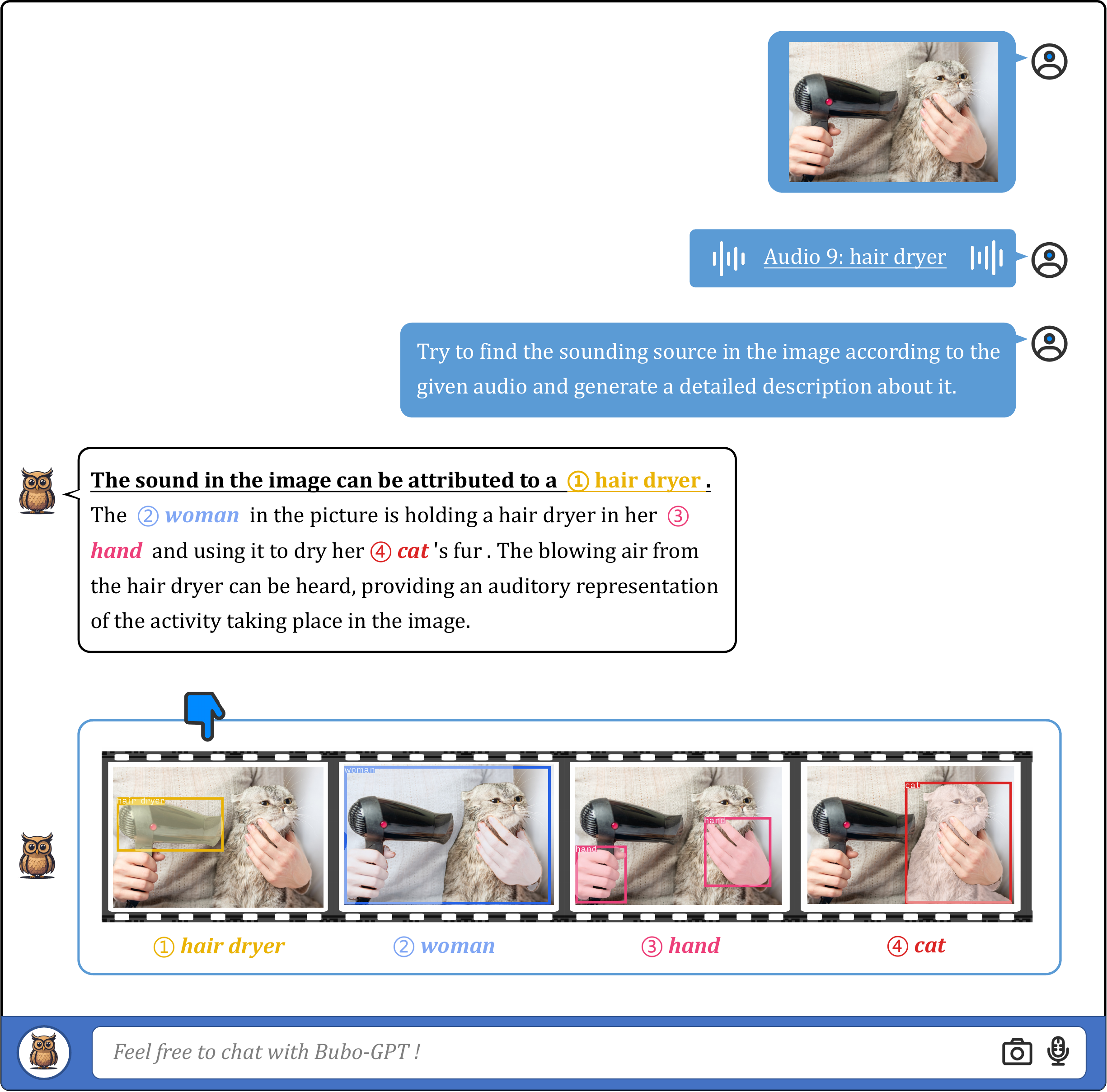}
  \caption{Example 3 of aligned audio-image understanding. Audio 9 can be accessed via this \href{https://huggingface.co/datasets/magicr/BuboGPT/blob/main/Audios/Audio-9-hair-dryer.mp3}{link}.}
  \label{figure:exp-pic-aud-3}
\end{figure*}

\begin{figure*}[t]
  \centering
  \includegraphics[width=\linewidth]{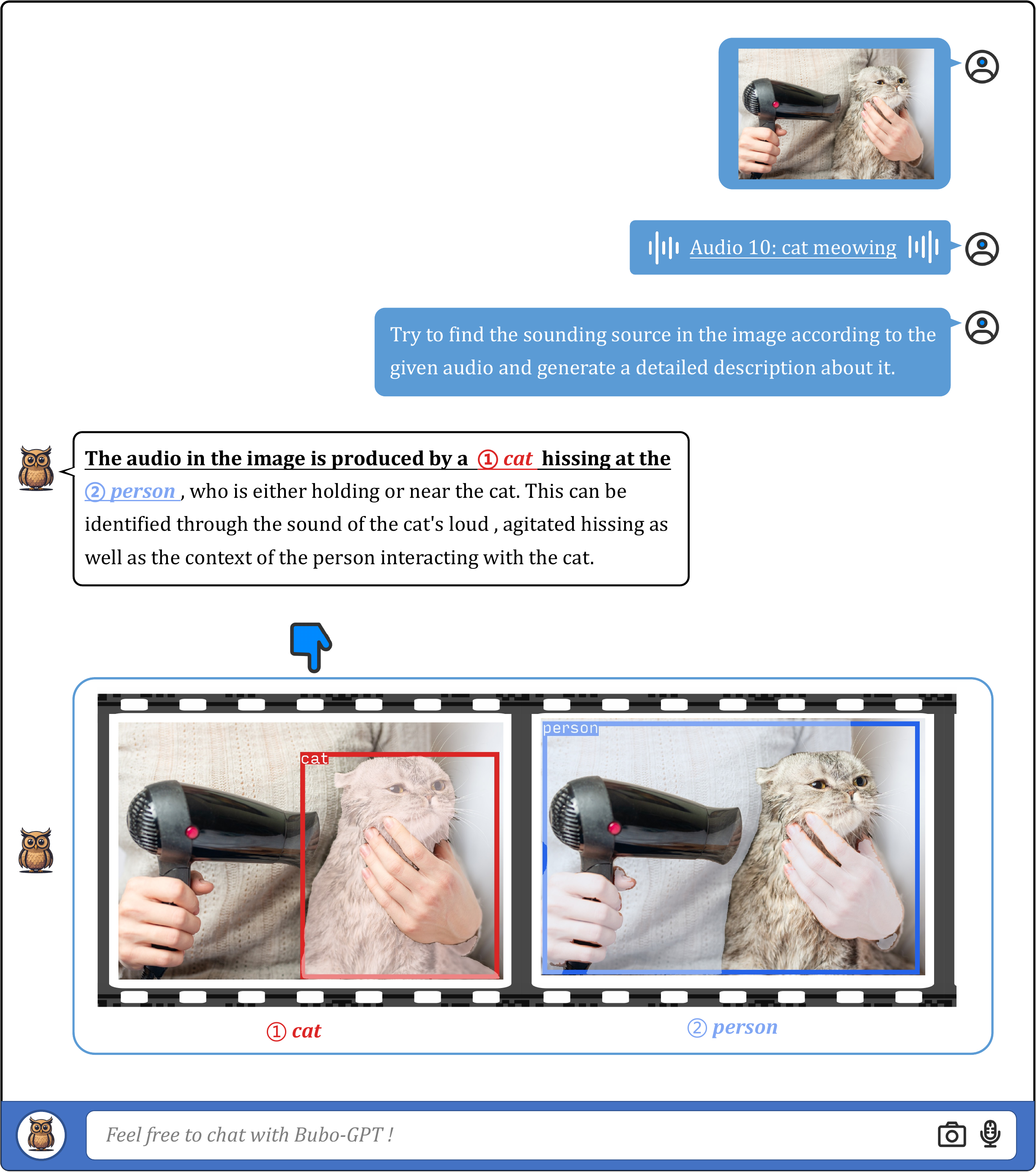}
  \caption{Example 4 of aligned audio-image understanding. Audio 10 can be accessed via this \href{https://huggingface.co/datasets/magicr/BuboGPT/blob/main/Audios/Audio-10-cat-meow.wav}{link}.}
  \label{figure:exp-pic-aud-4}
\end{figure*}

\begin{figure*}[t]
  \centering
  \includegraphics[width=\linewidth]{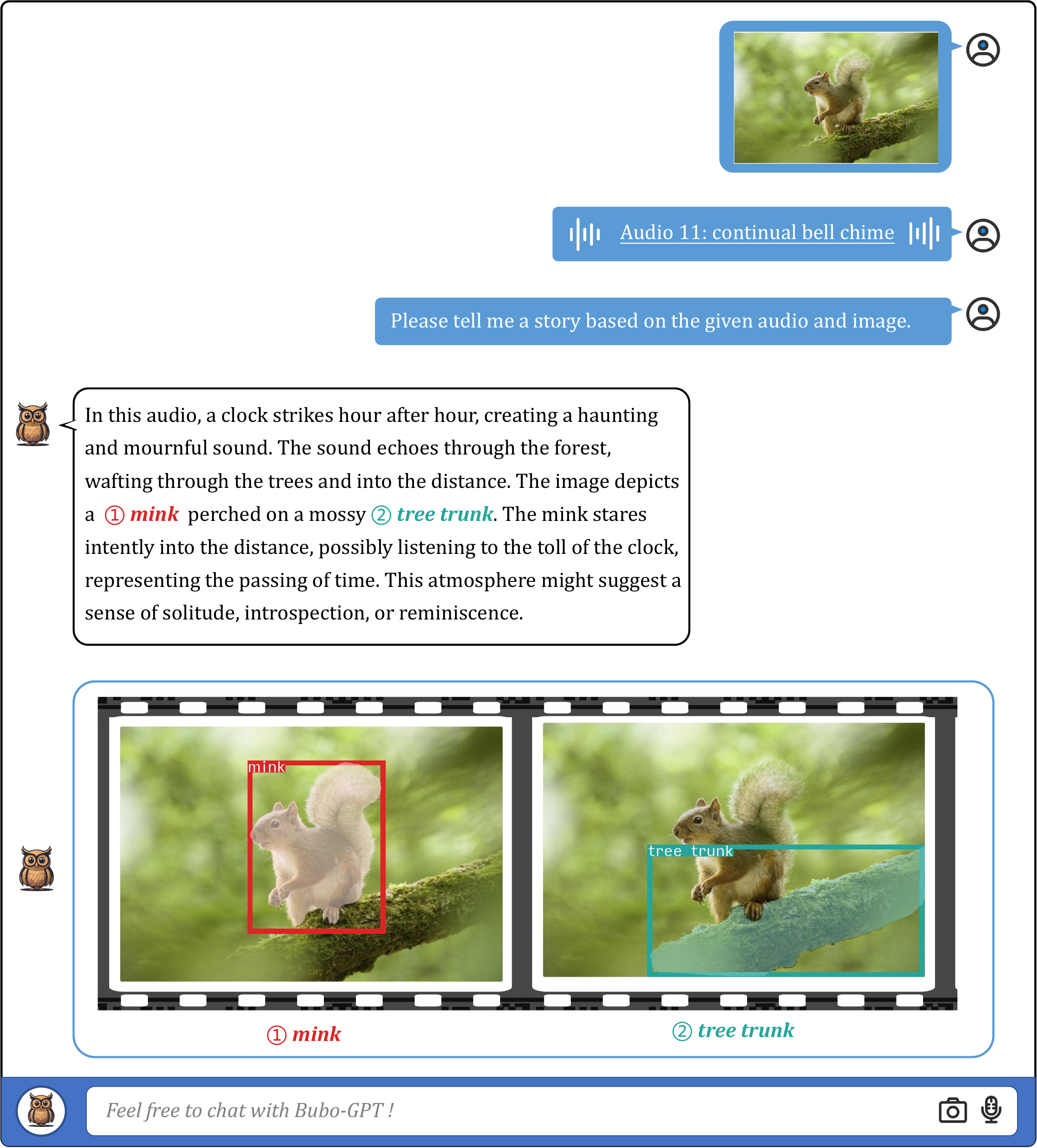}
  \caption{Example 1 of arbitrary audio-image understanding. Audio 11 can be accessed via this \href{https://huggingface.co/datasets/magicr/BuboGPT/blob/main/Audios/Audio-11-six-oclock.wav}{link}.}
  \label{figure:exp-pic-aud-5}
\end{figure*}

\begin{figure*}[t]
  \centering
  \includegraphics[width=\linewidth]{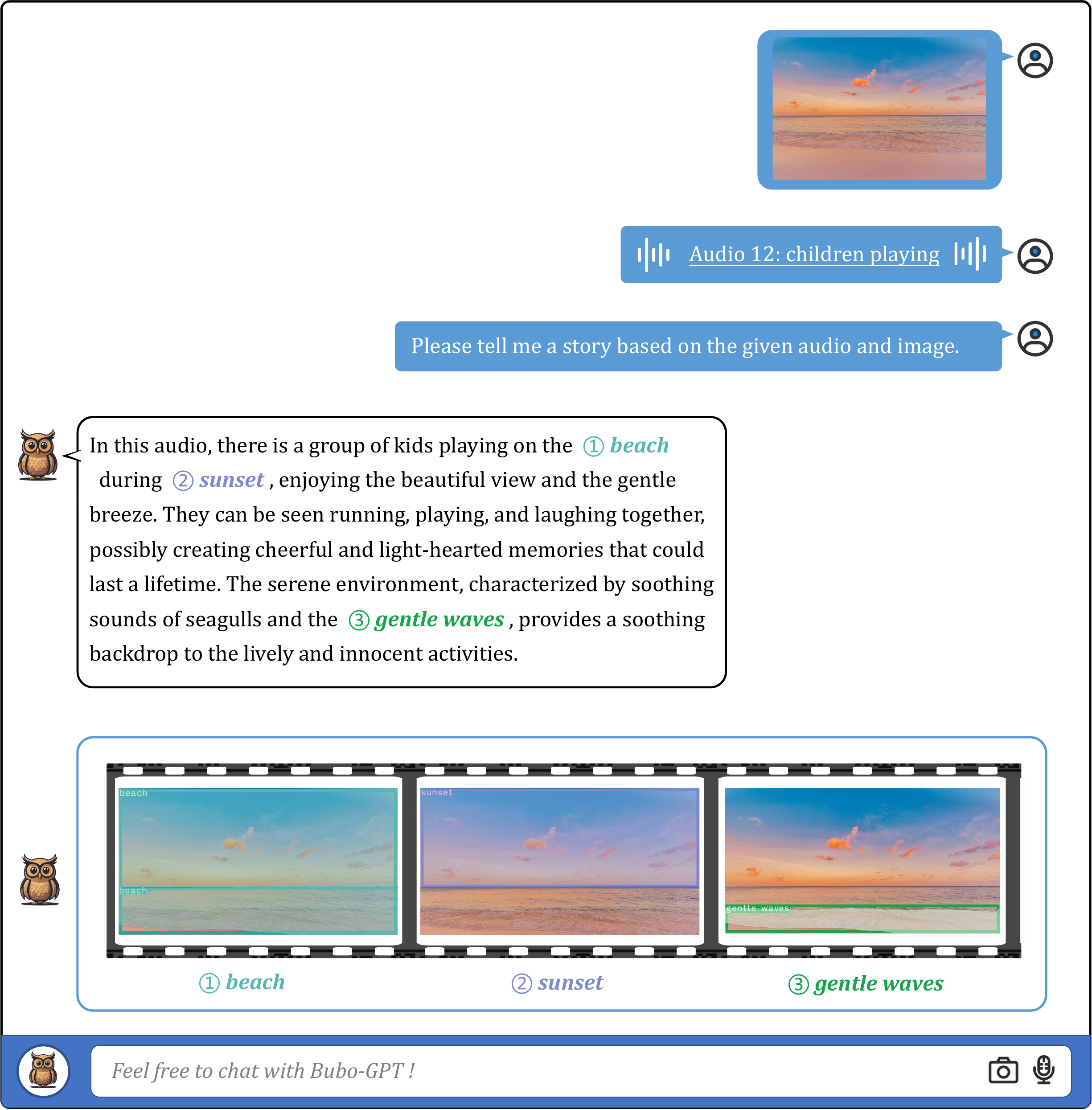}
  \caption{Example 2 of arbitrary audio-image understanding. Audio 12 can be accessed via this \href{https://huggingface.co/datasets/magicr/BuboGPT/blob/main/Audios/Audio-12-children-playing.mp3}{link}.}
  \label{figure:exp-pic-aud-6}
\end{figure*}

\begin{figure*}[t]
  \centering
  \includegraphics[width=\linewidth]{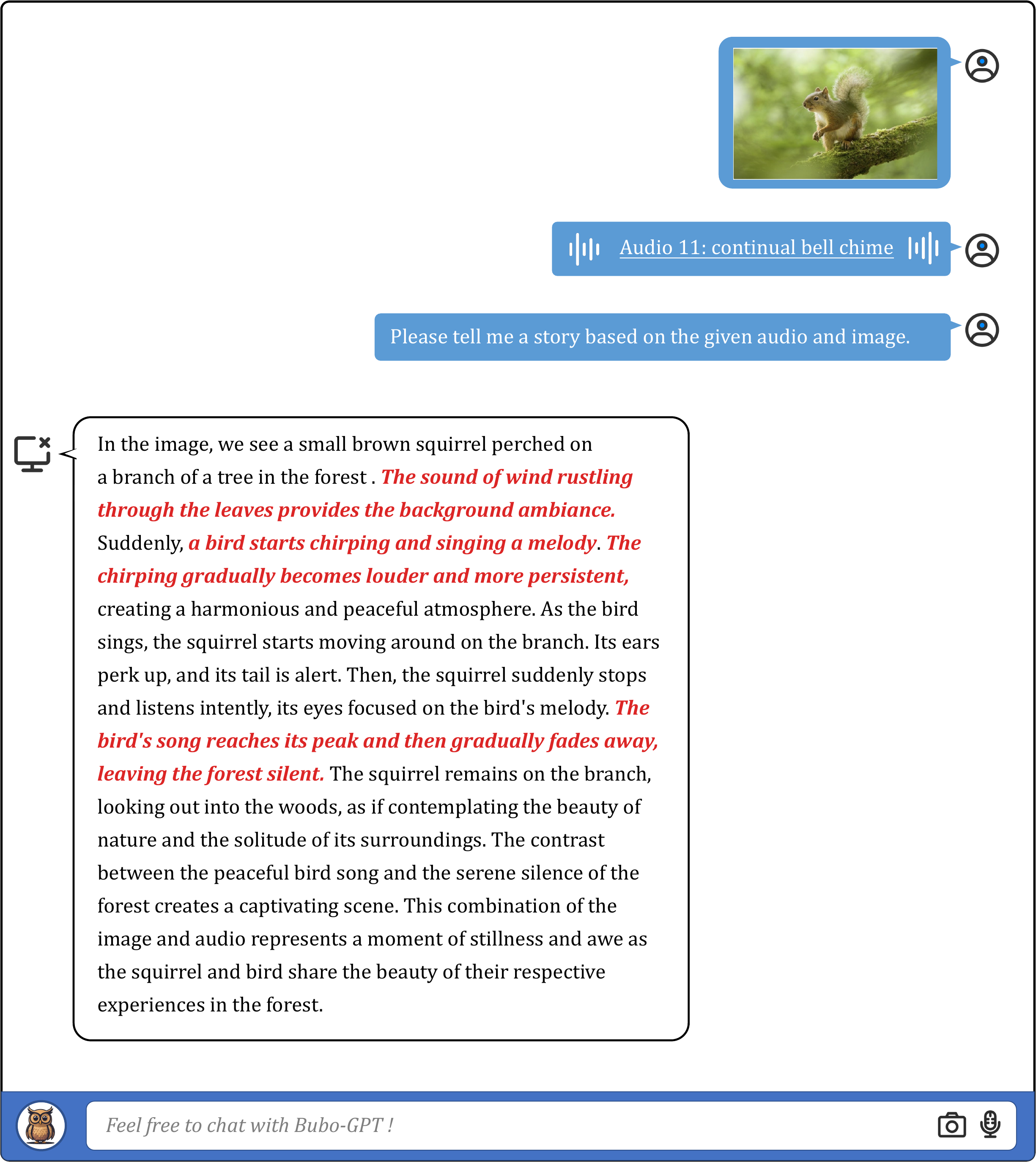}
  \caption{Failure case of arbitrary audio-image understanding without using negative audio-image pairs.}
  \label{figure:exp-pic-aud-error}
\end{figure*}

We first consider using a single image as input for \textbf{fine-grained visual understanding with grounding}. As shown in Fig.~\ref{figure:exp-pic-1}-\ref{figure:exp-pic-5}, the model can accurately associate textural words or phrases with image regions in various scenarios with different complexities. Then, when a single audio clip is provided for \textbf{audio understanding}, \name gives informative descriptions covering nearly all acoustic parts included, even when some audio fragments are too short for humans to notice, see Fig.~\ref{figure:exp-aud-1}-\ref{figure:exp-aud-6} for details. Next, we show that the model can perform sound localization with a matched audio-image pair provided, which gives a perfect example for \textbf{aligned audio-image understanding}. As illustrated in Fig.~\ref{figure:exp-pic-aud-1}-\ref{figure:exp-pic-aud-4}, the model is going to generate an overall description for both input image and audio, then point out which object in the image emits the sound after reasoning. It is worth noting that our model can give correct predictions when we provide different audio and keep the image unchanged. This demonstrates that our model can understand both modalities comprehensively rather than generate answers with prior bias from a single modality. Moreover, we empirically accessed that if the model is only tuned with well-aligned image-audio data, it actually fails to discriminate when an irrelevant image and audio pair is provided, resulting in a non-factual response that is not consistent with the given image or audio (\textcolor{blue}{Fig. ~\ref{figure:exp-pic-aud-error}}). After introducing the negatively paired dataset as discussed in Sec.~\ref{sec:audio-image-text-data}, the model can tell whether the image and audio are relevant to each other and generate high-quality response for \textbf{arbitrary audio-image understanding}, as evidenced by Fig.~\ref{figure:exp-pic-aud-5}-\ref{figure:exp-pic-aud-6}.

\section{Conclusion}
In this report, we propose a multi-modal LLM, \name, which is capable of joint multi-modal understanding including image, audio and text, and perform more fine-grained understanding of multi-modal
inputs by exploring the relation between different visual objects and modalities. We also build a high-quality instruction tuning dataset and the experiments show that \name achieves impressive visual grounding abilities during multi-modal chat, even when arbitrary combinations of multi-modal inputs are provided, whether matched or unmatched.  

\section{Limitations}
\textbf{Language hallucination.} Following prior works, our method is based on the pre-trained Vicuna model, which inherits the limitations of LLMs including generating non-existent knowledge or non-factual information. The problem might be resolved by training with more high-quality data and developing trustworthy LLMs.

\textbf{Inadequate capacities of Grounding QA. } Since the connection between grounding results and different modalities is built upon text conversations without extra training, the capacities of QA on specific objects remain limited. The model can be improved by introducing fine-grained visual grounding datasets and considering the spatial location as extra input.

\clearpage
{\small
\bibliographystyle{unsrt}
\bibliography{biblio}
}

\end{document}